\documentclass{article}

\usepackage{PRIMEarxiv}

\usepackage[utf8]{inputenc} 
\usepackage[T1]{fontenc}    
\usepackage{hyperref}       
\usepackage{url}            
\usepackage{booktabs}       
\usepackage{amsfonts}       
\usepackage{nicefrac}       
\usepackage{microtype}      
\usepackage{lipsum}
\usepackage{fancyhdr}       
\usepackage{graphicx}       
\graphicspath{{media/}}     
\usepackage{authblk}
\usepackage{hyperref}

\usepackage{orcidlink}
\usepackage{multirow}
\usepackage{booktabs, threeparttable}

\title{An Automated Classifier of Harmful Brain Activities for Clinical Usage Based on a Vision-Inspired Pre-trained Framework
}

\author[1]{Yulin Sun \orcidlink{0000-0003-3097-7755}}
\author[1]{Xiaopeng Si}
\author[1]{Runnan He}
\author[4]{Xiao Hu}
\author[5]{Peter Smielewski}
\author[1]{Wenlong Wang}
\author[6]{Xiaoguang Tong}
\author[6]{Wei Yue}
\author[1]{Meijun Pang}
\author[1]{Kuo Zhang}
\author[1]{Xizi Song}
\author[1, 2, 3]{Dong Ming}
\author[1, 1, 2, 3, 4]{Xiuyun Liu \thanks{Corresponding author: xiuyun\_liu@tju.edu.cn}}
\affil[1]{Medical School, Tianjin University, Tianjin, China}
\affil[2]{State Key Laboratory of Advanced Medical Materials and Devices, Tianjin University, Tianjin, China}
\affil[3]{Haihe Laboratory of Brain-computer Interaction and Human-machine Integration, Tianjin, China}
\affil[4]{Nell Hodgson Woodruff School of Nursing, Emory University, Atlanta, GA, United States of America}
\affil[5]{Brain Physics Laboratory, Division of Neurosurgery, Department of Clinical Neurosciences, University of Cambridge, Cambridge, UK}
\affil[6]{Neurology Department, Tianjin Huanhu Hospital, Tianjin, China}

\begin{document}
\maketitle

\begin{abstract}
\textbf{Importance} Timely identification of harmful brain activities (e.g., seizures, periodic discharges) via electroencephalography (EEG) is critical for brain disease diagnosis and treatment, which remains limited application due to inter-rater variability, resource constraints, and poor generalizability of existing artificial intelligence (AI) models.

\textbf{Objective} To develop an automated classifier, i.e. VIPEEGNet, leveraging the advantage of transfer learning from ImageNet-pretrained vision models to distinguish six harmful brain activities using multichannel one-dimensional EEG signals.

\textbf{Design, Setting, and Participants} In this study, a convolutional neural network model, VIPEEGNet, was developed and validated using EEGs recorded from Massachusetts General Hospital/Harvard Medical School. The VIPEEGNet was developed and validated using two independent datasets, collected between 2006 and 2020: the development cohort included EEG recordings from 1950 patients, with 106,800 EEG segments annotated by at least one experts (ranging from 1 to 28). The online testing cohort consisted of EEG segments from a subset of an additional 1,532 patients, each annotated by at least 10 experts.

\textbf{Main Outcomes and Measures} The primary outcomes were the sensitivity, precision, and area under the receiver operating characteristic curve (AUROC) compared with the consensus voted by experts for six EEG patterns, including seizure, generalized periodic discharges (GPD), lateralized periodic discharges (LPD), generalized rhythmic delta activity (GRDA), and lateralized rhythmic delta activity (LRDA), and “other”. The more refined outcomes are the Kullback-Leibler Divergence (KLD) between the probability distributions of annotations labeled by experts and the probability predicted by the model for each category.

\textbf{Results} For the development cohort (n=1950), the VIPEEGNet achieved high accuracy, with an AUROC for binary classification of seizure, LPD, GPD, LRDA, GRDA, and “other” categories at 0.972 (95\% CI, 0.957-0.988), 0.962 (95\% CI, 0.954-0.970), 0.972 (95\% CI, 0.960-0.984), 0.938 (95\% CI, 0.917-0.959), 0.949 (95\% CI, 0.941-0.957), and 0.930 (95\% CI, 0.926-0.935). For multi classification, the sensitivity of VIPEEGNET for the six categories ranges from 36.8\% to 88.2\% and the precision ranges from 55.6\% to 80.4\%, and performance similar to human experts. Notably, the external validation showed KLDs of 0.223 and 0.273, ranking top 2 among the existing 2,767 competing algorithms, while we only used 2.8\% of the parameters of the first-ranked algorithm.

\textbf{Conclusions and Relevance} VIPEEGNet demonstrates that transfer learning from vision models enables efficient, clinically applicable EEG classification. Its minimal parameter requirements and modular design offer a deployable solution for real-time brain monitoring, potentially expanding access to expert-level EEG interpretation in resource-limited settings. The model source code can be accessed via Kaggle (\url{https://www.kaggle.com/code/sunyuri/hms-vipeegnet})

\end{abstract}


\section{Introduction}
Globally, an estimate of 43\% of the world population suffers from neurological diseases, becoming the leading cause of overall disease burden in the world\cite{steinmetz2024global}. As the most often used tool, the electroencephalography (EEG) provides essential information for brain disease diagnosis and guidance of the treatment\cite{tveit2023automated}. Timely identification of harmful brain activity via EEG plays critical role in managing epilepsy patients and critically ill individuals, underpinning crucial decisions in seizure detection, brain deterioration monitoring, treatment outcome prediction and rehabilitation planning\cite{bitar2024utility}. Evidence further suggests that other epileptiform patterns, such as generalized periodic discharges (GPD), lateralized periodic discharges (LPD), generalized rhythmic delta activity (GRDA), and lateralized rhythmic delta activity (LRDA), etc., the abnormal neuronal activities between overt seizures and normal EEG may also cause neuronal damage\cite{ge2021deep}. While these patterns provide a conceptual framework for understanding pathology, we still lack an accurate classifier to distinguish different abnormal brain activities, especially those harmful ones for early interventions.

\begin{figure}[htbp]
    \centering
    \includegraphics[width=0.98\textwidth]{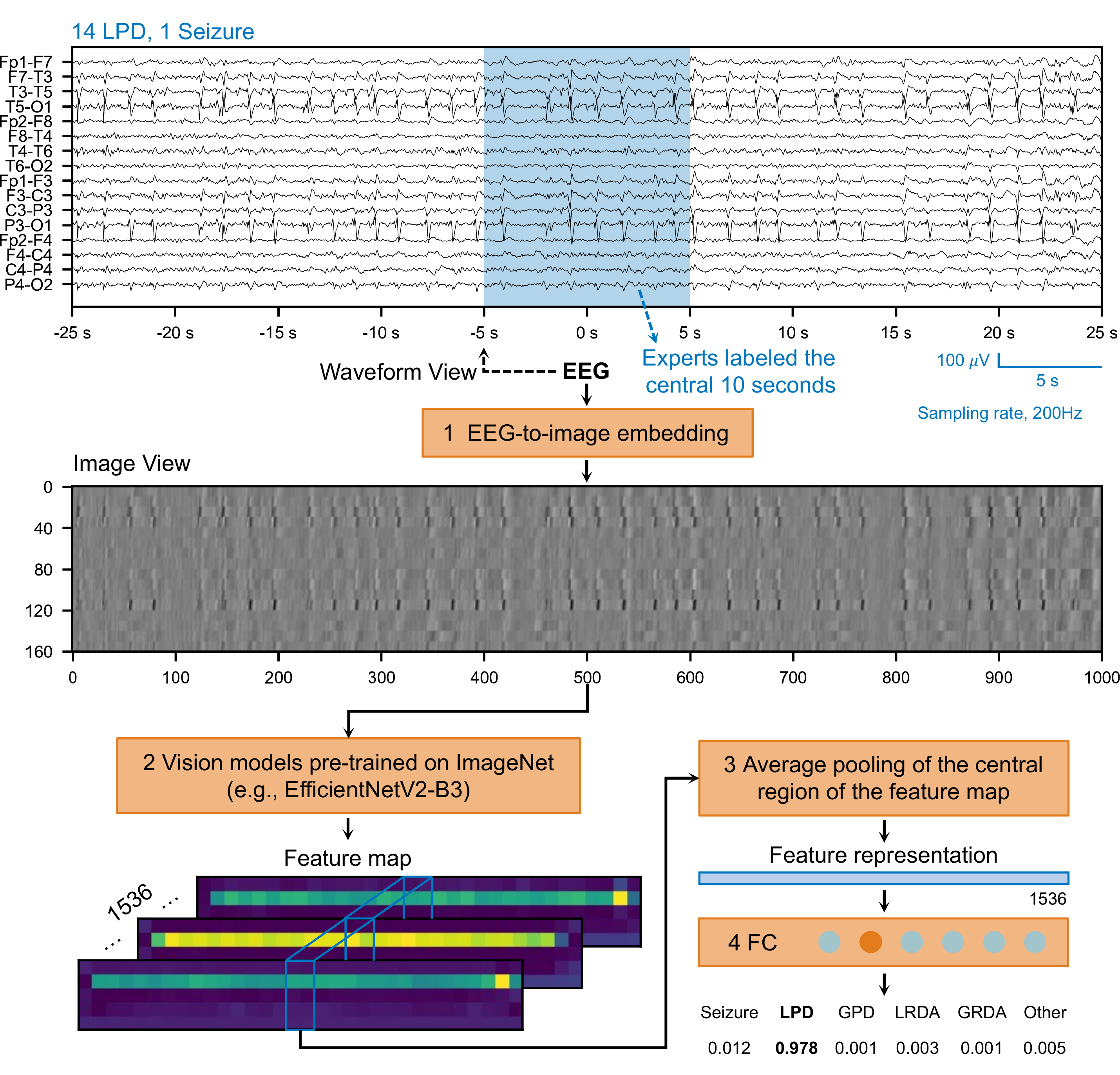}
    \caption{\textbf{The schematic diagram of the VIPEEGNet model designed for harmful brain activity classification.} The model takes 16-channel scalp EEG signals with a double banana reference as input. The EEG-to-image component converts the EEG signals into image representations. These image representations are then input into a pre-trained computer vision classification backbone for feature extraction. Average pooling is applied to the central regions of the temporal dimension of the feature maps. Finally, the feature representation is input into a fully connected (FC) layer to obtain the probabilities of different EEG patterns.}
    \label{fig:fig1}
\end{figure}

The current gold standard of manual EEG analysis by specialized neurologists presents significant limitations, including being highly time-consuming, costly, indistinguishable to fatigue-induced errors, and poor assessment consistency\cite{jing2023interrater}. Although advances in deep learning and the availability of larger EEG datasets have spurred the development of automated algorithms for classifying seizures and related patterns\cite{saab2024towards, hogan2025scaling}, critical obstacles hinder their practical application and generalization. Most existing models are trained and validated on relatively small datasets, demonstrating poor performance in handling the inherent variability of EEG signals across diverse patient populations\cite{sun2022continuous, sun2025multi}. Even more critically, most models utilize one-dimensional EEG signals as input, lacking the robust and versatile pre-trained models that are prevalent in the field of computer vision, such as ResNet, ViT, Swin Transformer trained on ImageNet\cite{krizhevsky2017imagenet, russakovsky2015imagenet}. Consequently, the existing models optimized for specific datasets show poor transferability for a new clinical data set, and is difficult to achieve a balance between high accuracy, model efficiency, ease of deployment and broad generalizability. This challenge raises a pivotal question: Can the advances of transfer learning from state-of-the-art ImageNet-pretrained vision models, which is dominantly used in computer vision field for imaging classification, be effectively utilized to distinguish harmful brain activities by using multichannel one-dimensional EEG signals? Our goal was to propose an AI model, i.e. VIPEEGNet (Vision-Inspired Pre-trained Network for EEG Classification, shown in Figure \ref{fig:fig1}), for automatic classification of six critical brain activity patterns, i.e. seizure (SZ), GPD, LPD, LRDA, GRDA, and “other”\cite{hirsch2021american}. Our core innovation is a general EEG-to-image conversion module, enabling the direct adaptation of state-of-the-art (SOTA) ImageNet pre-trained vision models for EEG analysis. This strategy, using only 2.8\% of parameters, matches the excellent performance of complex algorithms that use 30 backbone networks as feature extractors, achieving significant efficiency.

We trained a deep learning model on a large data set of highly annotated EEG recordings from 1950 subjects (each EEG record was meticulously annotated by at least 3 experts and conducted external online testing. The new established model, which we made publicly available on Kaggle (\url{https://www.kaggle.com/code/sunyuri/hms-vipeegnet}), exhibits significant generality and features a future-oriented modular design. The findings prove that leveraging powerful pre-trained vision models based on ImageNet is able to achieve excellent EEG classification performance. VIPEEGNet represents a significant step towards bridging computer vision and EEG analysis, demonstrating the feasibility and power of transfer learning for this critical medical task and opening a new path for efficient and accurate EEG classification at clinical bedside.

\section{Methods}
\subsection{Datasets}
The datasets used in this study is available in Kaggle and can be accessed via this link \url{https://kaggle.com/competitions/hms-harmful-brain-activity-classification}. It was developed by Dr. Westover and his team members from the Massachusetts General Hospital/Harvard Medical School Department of Neurology\cite{jing2023interrater, jing2023development, barnett2024improving}. A portion of this database has been uploaded to the Kaggle platform for researchers to develop and test their algorithms\cite{Jing2024hms}. The data can be divided into two cohorts: the development cohort and the online testing cohort.

The development cohort was used for algorithm training and validation and included EEG recordings from 1950 patients (mean age 47.73 years [SD 25.52]; 963 females [49.38\%]), with experts annotating 106,800 EEG segments (Table \ref{tab:tab1}). The EEG segments corresponding to seizure, LPD, GPD, LRDA, GRDA, and “other” events were 20,933, 14,856, 16,702, 16,640, 18,861, and 18,808. It can be divided into two datasets based on the number of experts involved in the annotation: Dataset 1, composed of EEG recordings from 1794 patients, with 11900 EEG recordings and 68854 EEG segments, annotated by 1-7 experts, and Dataset 2 (referred to as the high-quality annotation subset in this paper), composed of EEG recordings from 1,063 patients, with 5,939 EEG recordings and 39,946 EEG segments, annotated by 10-28 experts.

\begin{table}[htbp]
  \centering
  \caption{The EEG segments and expert annotations of the development dataset.}  
  \label{tab:tab1}
  \begin{tabular}{llll}
  \toprule
  ~ & Whole & {Subset 1, 1-7 labels} & {Subset 2, 10-28 labels} \\ 
  \hline
  Patient Number & 1950 & 1794 & 1063 \\ 
  {Age, y, mean (SD)} & {47.73 (25.52)} & - & - \\ 
  {Female, n (\%)} & {963 (49.38)} & - & - \\ 
  \hline
  Segments, n (\%) & 106,800 (100) & 68,854 (100) & 39,946 \\
  Seizure	& 20,933 (19.6) & 20,337 (29.54) & 596 (1.49) \\
  LPD	& 14,856 (13.91) & 7,416 (10.77) & 7,440 (18.63) \\
  GPD	& 16,702 (15.64) & 6,625 (9.62) & 10,077 (25.23) \\
  LRDA & 16,640 (15.58) & 10,410 (15.12) & 6,230 (15.6) \\
  GRDA & 18,861 (17.66) & 13,478 (19.57) & 5,383 (13.48) \\
  Other & 18,808 (17.61) & 8,588 (12.47) & 10,220 (25.58) \\
  \hline
  EEG recordings, n (\%)	& 17,089 (100)	& 11,900 (100)	& 5,939 (100) \\
  Seizure	& 3,973 (23.25)	& 3,694 (31.04)	& 279 (4.7) \\
  LPD	& 3,350 (19.6)	& 2,055 (17.27)	& 1,295 (21.81) \\
  GPD	& 2,081 (12.18)	& 806 (6.77)	& 1,275 (21.47) \\
  LRDA	& 1,132 (6.62)	& 771 (6.48)	& 361 (6.08) \\
  GRDA	& 1,930 (11.29)	& 1,483 (12.46)	& 447 (7.53) \\
  Other	& 7,717 (45.16)	& 4,555 (38.28)	& 3,162 (53.24) \\
  \bottomrule
  \end{tabular}
\end{table}

EEG electrodes were placed according to the 10-20 placement standard, and EEG data were resampled to a rate of 200 Hz. Each EEG segment was 50 seconds long, and experts annotated the event type for the middle 10 seconds. EEG data were re-referenced using longitudinal bipolar/double banana methods. Experts could refer not only to the 50-second EEG but also to the spectrogram of the preceding and following 5 minutes as auxiliary information for annotation.

The online testing cohort consisted of a subset of EEG segments from an additional 1,532 patients, with each EEG sample annotated by at least 10 experts. Researchers were only allowed to upload classification models to classify the test data, and the website returned the model performance evaluation results without allowing direct access to the test set to strictly prevent data leakage.

\subsection{Training, validation, and testing}
\begin{figure}[htbp]
    \centering
    \includegraphics[width=0.98\textwidth]{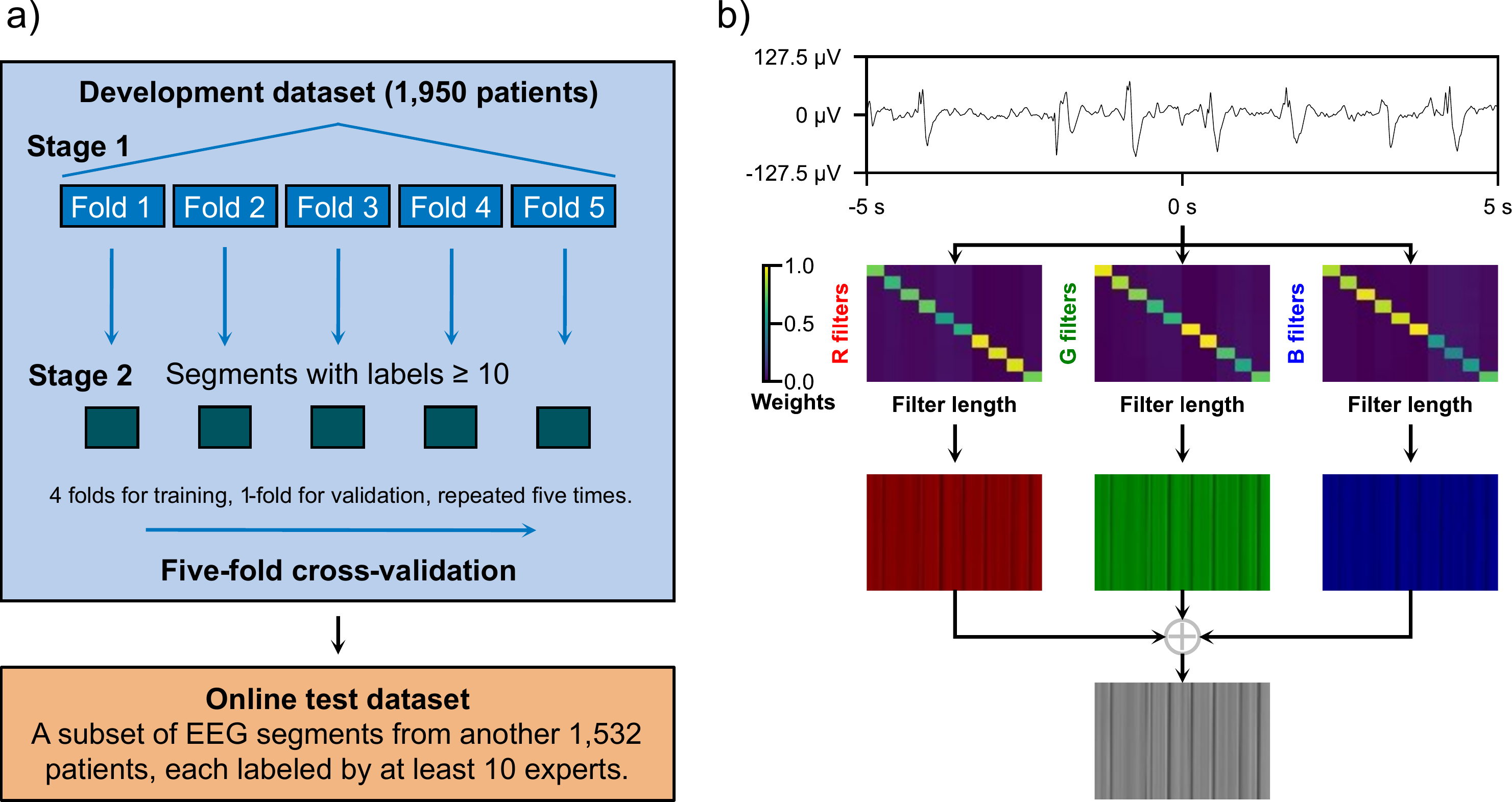}
    \caption{\textbf{Training, validation, and testing strategies, and details of the EEG-to-image module.} (a) Five-fold cross-validation was used to evaluate the model's performance on the development cohort. Each fold involved two-stage training and validation. The optimal model determined by the validation set was used for final assessment on the online test dataset. (b) Details of the EEG-to-image component. EEG signals are transformed into image representations via three groups of learnable convolutional kernels based on the red, green, and blue color theory.}
    \label{fig:fig2}
\end{figure}

In the development cohort, a five-fold cross-validation strategy was employed to reliably assess model performance. The dataset was split into five folds, ensuring that samples from the same patient remained in the same fold to prevent data leakage. As shown in Figure \ref{fig:fig2}a, for each fold, the model was trained in two stages. In the first stage, the model was trained on the entire dataset, with the number of experts' annotations for each sample used as weights in the loss function to balance the knowledge contribution of the experts. The learning rate was set to 0.001, and training lasted for 15 epochs. In the second stage, the model was fine-tuned on a subset of data from the current fold (samples with 10+ labels), with each sample assigned the same weight in the loss function. The learning rate was reduced to 0.0003, and the training duration was adjusted to 5 epochs. This two-stage training approach allowed the model to first learn general features and then improve its predictions on a more reliable data subset.

\subsection{Model architecture}
The architecture of VIPEEGNet is shown in Figure Figure \ref{fig:fig1}. The model takes EEG data as input, defined by the number of channels and signal length. As depicted in Figure \ref{fig:fig2}b, a 1D convolutional layer is used to embed EEG signals into image representations. The EEG-to-image component is implemented with three groups of convolutional kernels corresponding to the R, G, and B color channels. Taking the R group as an example, it consists of 10 convolutional kernels of length 10 with a stride of 10. This layer is initialized using a custom initializer and constrained to ensure that the weights of each convolutional kernel are positive and sum to 1, promoting meaningful feature extraction. The EEG-to-image component reduces the time scale of the original EEG signal to one-tenth of its original size while expanding the spatial scale tenfold. This can be seen as a learnable down sampling process that preserves the overall information content. In the EEG-to-image embedding process, adjacent sampling points in the time domain are transformed into adjacent points in the spatial domain, resulting in a transformed image where the horizontal axis contains temporal information and the vertical axis contains spatiotemporal information, homogenizing the attributes of the two axes.

The EEG-to-image component reduces the time scale of the original EEG signal to one-tenth of its original size while expanding the spatial scale tenfold. This can be seen as a learnable down sampling process that preserves the overall information content. Subsequently, the EfficientNetV2-B3 model, loaded with pre-trained weights to leverage transfer learning, is used as the feature extraction backbone\cite{tan2104efficientnetv2}. After feature extraction with EfficientNet, the model retains the intermediate region of the feature map along the time scale to match the experts' annotation practice (observing a 50-second sample and annotating the event type for the middle 10 seconds). Global average pooling is then applied to reduce feature dimensions, followed by a dropout layer to prevent overfitting. Finally, a dense layer with SoftMax activation generates classification probabilities for different brain activities.

\subsection{Evaluating performance}
During training, the model's performance was evaluated using the KLD loss function. Confusion matrices were generated for each fold and stage to visualize the model's classification performance across different brain activity categories. Training and validation loss curves were plotted to monitor model convergence. To ensure reproducibility, random seeds were set for relevant functions, and deterministic operations were enabled in TensorFlow. EEG signals were bandpass filtered using a third-order Butterworth filter between 0.5 Hz and 45 Hz to remove noise and artifacts outside the frequency range of interest. The EEG signals were then clipped to the range of -1024 µV to 1024 µV to handle outliers and scaled to 0-255 for standardization to suit the input range of image-based pre-trained models. During training, various data augmentation techniques were applied to the EEG signals to enhance the model's robustness and generalization ability. These techniques included random masking of portions of the EEG signal, channel permutation, signal inversion, time reversal, and channel swapping.

A custom cosine annealing learning rate scheduler was implemented to dynamically adjust the learning rate during training. This scheduler combined a warm-up phase with a cosine annealing strategy to optimize convergence. During training, model checkpoints were regularly saved based on validation loss, and the best-performing model weights were retained for subsequent evaluation. For online testing, predictions from models trained on different folds were averaged to generate the final classification probabilities for each EEG sample. These probabilities were then saved in the required format for submission.

\section{Results}
\subsection{Five-fold cross-validation performance}
\begin{figure}[htbp]
    \centering
    \includegraphics[width=0.96\textwidth]{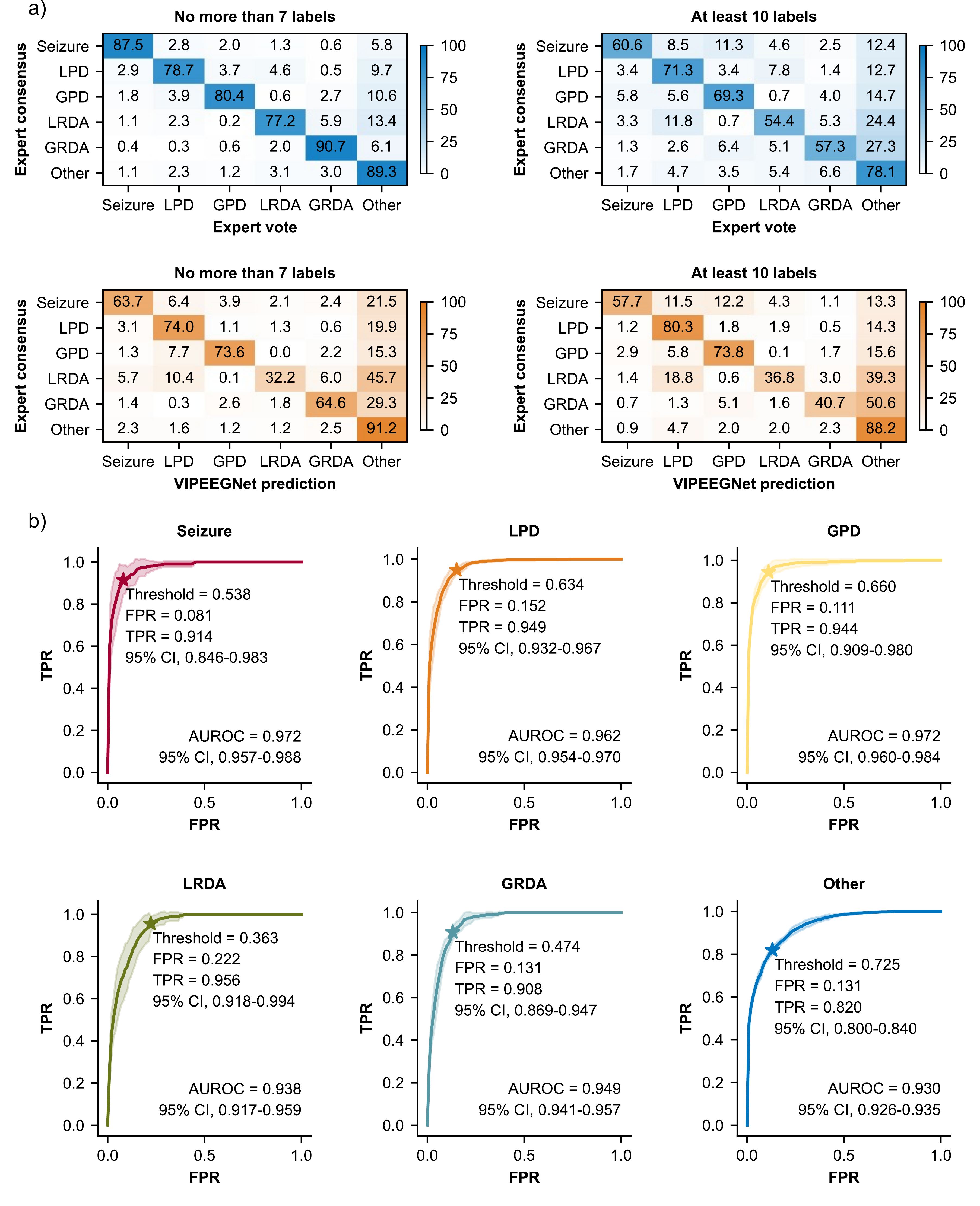}
    \caption{\textbf{Five-fold cross-validation results.} (a) Confusion matrices of VIPEEGNet predictions versus expert annotations. Confusion matrices computed with expert majority consensus as the reference. The first column corresponds to the minority-expert-labeled dataset, and the second column corresponds to the high-quality-labeled dataset. (b) One-vs-rest ROC curves for six-class classification. Each curve represents the classification performance for one class against the other five, providing a comprehensive view of the model's ability to distinguish each class from the rest. The pentagram represents the threshold corresponding to the optimal performance of the binary classification. AUROC: area under the receiver operating characteristic curve.}
    \label{fig:fig3}
\end{figure}

The development cohort used for algorithm training and validation, includes Dataset 1 (n=1794) and Dataset 2(n=1063). For each EEG segment, the event type with the highest frequency of annotations among the relevant expert annotations was determined as the expert consensus. Consequently, for each EEG segment, there may exist individual expert annotations that deviate from the established consensus. To facilitate the analysis of these discrepancies, we constructed a confusion matrix (Figure \ref{fig:fig3}a). In the confusion matrix, each row corresponds to an expert consensus class, and each column corresponds to the predicted classes by individual expert votes. This matrix serves as a comprehensive tool for visualizing and quantifying the patterns of disagreement between individual expert annotations and the collective consensus. In general, the sensitivity and precision of VIPEEGNET for the six categories were similar to human experts. For Dataset 1 (1,794 patients, each sample annotated by seven or fewer experts), the sensitivity and precision of expert voting for seizure, LPD, GPD, LRDA, GRDA, and “other” categories were (87.5, 96.9) \%, (78.7, 85.0) \%, (80.4, 79.4) \%, (77.2, 66.4) \%, (90.7, 85.7) \%, and (89.3, 81.0) \%, respectively. In comparison, VIPEEGNet achieved sensitivity and precision of (63.7, 90.5) \%, (74.0, 75.5) \%, (73.6, 78.3) \%, (32.2, 56.9) \%, (64.6, 76.8) \%, and (91.2, 65.5) \%, respectively. For Dataset 2 (1,063 patients, each sample annotated by ten or more experts), the sensitivity and precision of expert voting for seizure, LPD, GPD, LRDA, GRDA, and “other” categories were (60.6, 51.8) \%, (71.3, 76.6) \%, (69.3, 71.2) \%, (54.4, 37.0) \%, (57.3, 45.8) \%, and (78.1, 83.3) \%, respectively. VIPEEGNet demonstrated sensitivity and precision of (57.7, 68.2) \%, (80.3, 77.5) \%, (73.8, 80.4) \%, (36.8, 55.6) \%, (40.7, 63.0) \%, and (88.2, 79.6) \%, respectively. In this data set, the model’s precision was higher than human experts, indicating the model’s learning ability by using high-quality annotated data. Therefore, in the subsequent analyses of the model's five-fold cross-validation performance, Dataset 2 with the high-quality annotation was used.

The VIPEEGNet achieved high accuracy, with an area under the receiver operating characteristic curve (AUROC) for binary classification of seizure, LPD, GPD, LRDA, GRDA, and “other” categories at 0.972 (95\% CI, 0.957-0.988), 0.962 (95\% CI, 0.954-0.970), 0.972 (95\% CI, 0.960-0.984), 0.938 (95\% CI, 0.917-0.959), 0.949 (95\% CI, 0.941-0.957), and 0.930 (95\% CI, 0.926-0.935), as shown in Figure \ref{fig:fig3}b. The optimal classification thresholds were determined to be 0.538, 0.634, 0.660, 0.363, 0.474, and 0.725. It was observed that the model tended to misclassify EEG segments annotated as LRDA by expert consensus as other categories.

Furthermore, when considering the annotation results of multiple experts, traditional binary classification metrics are insufficient to describe the annotation of difficult samples. For instance, when ten experts annotate a sample as seizure and another ten as GPD. To address this limitation, the expert annotation results were normalized by dividing by the total number of experts annotating the sample to obtain soft labels. The Kullback-Leibler Divergence (KLD) was employed to calculate the similarity between the model's predicted probabilities and the expert annotations. The average KLD across five-fold cross-validation in the development dataset was 0.225 (95\% CI, 0.211-0.238). This metric provides a more nuanced evaluation of the model's performance, taking into account the variability in expert annotations.

\begin{table}[htbp]
  \centering
  \caption{The model's performance on the online test cohort compared with the top 5 gold medal algorithms out of 2767.} 
  \label{tab:tab2}
  \begin{tabular}{llllll}
  \toprule
  \multirow{2}{*}{Teams (2767)} & \multicolumn{2}{l}{Leaderboard (KLD)} & \multirow{2}{*}{View} & \multirow{2}{*}{Backbones}& \multirow{2}{*}{Parameters (M)} \\ 
  \cline{2-3}
  ~ & Public & Private & ~ & ~ & ~ \\ 
  \hline
  \multirow{2}{*}{Team Sony (1)} & \multirow{2}{*}{0.2161} & \multirow{2}{*}{0.2723} & EEG & {ConvNeXt Atto × 4 + InceptionNeXt Tiny} & \multirow{2}{*}{1762} \\
  ~ & ~ & ~ & Spectrogram & {MaxViT\_base × 11 + SwinV2-Tiny × 14} & ~ \\
  VIPEEGNet (ours) & 0.2233 & 0.2725 & EEG & EfficientNetV2-B3 & 49 \\
  \multirow{3}{*}{Holiday (2)} & \multirow{3}{*}{0.2243} & \multirow{3}{*}{0.2738} & EEG & {EfficientNet-B5 × 2 + PPHGNetV2\_B5} & \multirow{3}{*}{283} \\
  ~ & ~ & ~ & Spectrogram & {X3D-L + EfficientNet-B5} & ~ \\
  ~ & ~ & ~ & Multi-view & {X3D-L + EfficientNet-B5} & ~ \\
  \multirow{2}{*}{Nvidia/DD (3)} & \multirow{2}{*}{0.2255} & \multirow{2}{*}{0.2741} & EEG & {(1D CNN \& SqueezeFormer) × 5} & \multirow{2}{*}{274} \\
  ~ & ~ & ~ & Spectrogram & {(mixnet\_l or mixnet\_xl) × 18} & ~ \\
  \multirow{3}{*}{Aillis \& GO \& bilzard (4)} & \multirow{3}{*}{0.2186} & \multirow{3}{*}{0.2760} & EEG & {1D CNN + MobileNetV2} & \multirow{3}{*}{929} \\
  ~ & ~ & ~ & Spectrogram & {SwinV2-Large × 4} & ~ \\
  ~ & ~ & ~ & Multi-view & {Vit-Tiny + CAFormer + ConvNext} & ~ \\
  KTMUD (5) & 0.2259 & 0.2762 & Multi-view & WaveNet + 2D CNN & - \\
  \hline
  Silver medal (16-138) & ~ & ~ & ~ & ~ & ~ \\
  Mikhail Kotyushev (16) & 0.2377 & 0.2946 & Multi-view & Vit-Tiny & 21 \\
  \hline
  bronze medal (139-276) & ~ & ~ & ~ & ~ & ~ \\
  Yuki Take93 (139) & 0.2849 & 0.3534 & - & - & - \\
  \bottomrule
  \end{tabular}
  \begin{tablenotes}
    \small
    \item[] Abbreviations: KLD, Kullback-Leibler Divergence.
    \item[] For the details of algorithms, refer to \url{https://www.kaggle.com/competitions/hms-harmful-brain-activity-classification/leaderboard}.
  \end{tablenotes}
\end{table}

\subsection{Online testing results}
After demonstrating promising performance in the development datasets, the VIPEEGNet model was tested in an additional online extra cohort (a subset of an additional 1,532 patients). The online testing dataset was hosted on the Kaggle platform and made available in a competition format to 3,507 participants from 113 countries, forming 2,767 teams\cite{Jing2024hms}. The public leaderboard and private leaderboard used 35\% and 65\% of the testing data, respectively. VIPEEGNet achieved a KLD score of 0.223296 on the public leaderboard and 0.272543 on the private leaderboard (Table \ref{tab:tab2}). Its overall performance ranked second among 2,767 algorithms, trailing the top algorithm by only 0.0002 in KLD score. Notably, compared to the SOTA algorithm based on raw EEG and spectrogram using 30 model backbones, VIPEEGNet used only a single EEG-based model, reducing the parameter count to 2.8\% of the former and offering higher transparency and interpretability, which is beneficial for practical deployment.

\subsection{Ablation experiments}
\begin{figure}[htbp]
    \centering
    \includegraphics[width=0.98\textwidth]{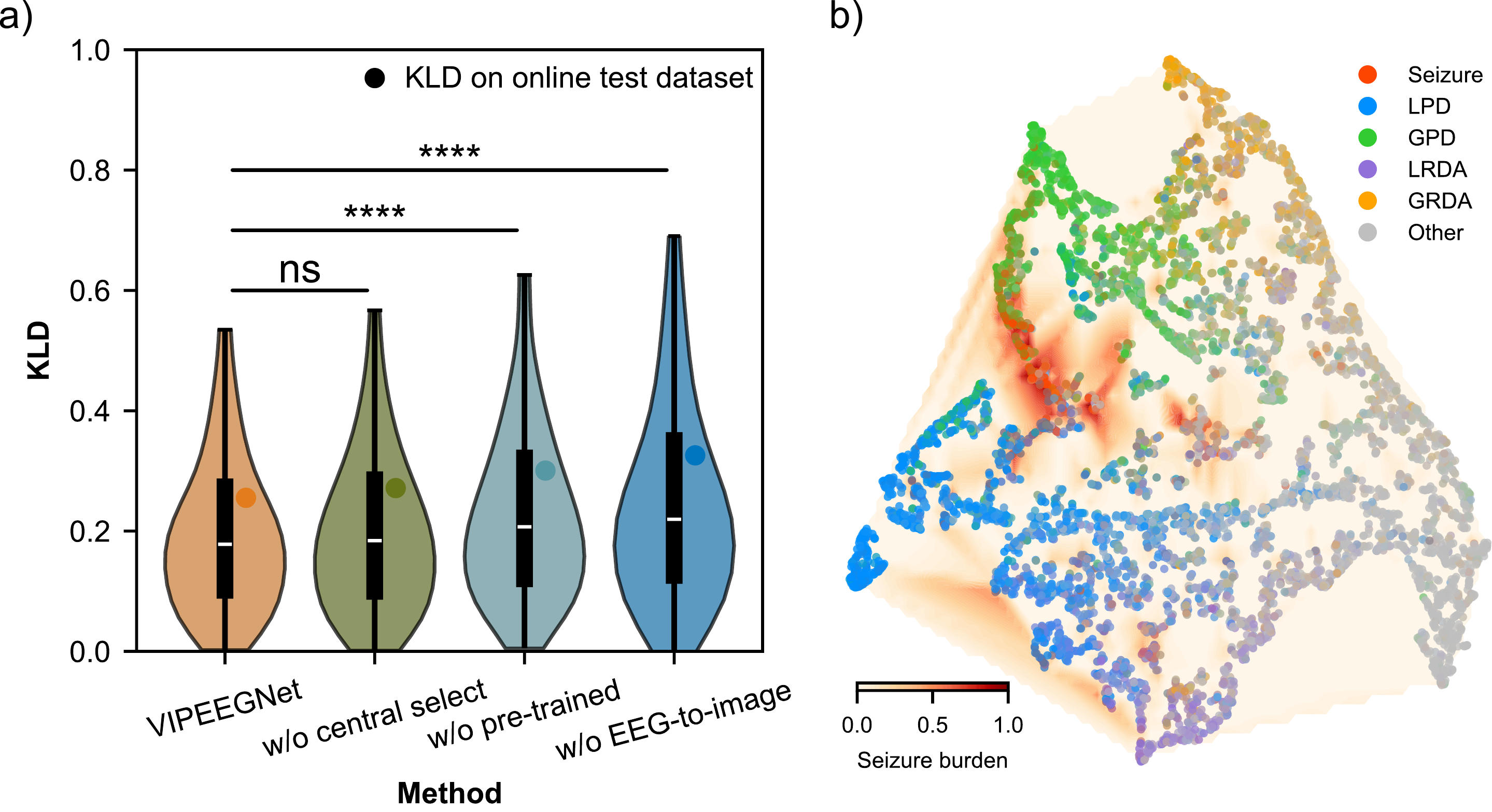}
    \caption{\textbf{Model interpretability analysis.} (a) Ablation studies comparing model performance by removing different components (central selection, pretraining, EEG-to-image). (b) t-SNE visualization of model outputs with dimensional reduction, where colors represent expert consensus.}
    \label{fig:fig4}
\end{figure}

The ablation experiments provided valuable insights into the contribution of each component of VIPEEGNet. The results, as shown in Figure \ref{fig:fig4}a, indicated that the removal of the EEG-to-image component had the most significant impact on model performance (0.275 vs. 0.216, p < 0.0001), and the removal of the pre-training component also led to a significant decline in performance (0.258 vs. 0.216, p < 0.0001). These findings demonstrate that converting EEG into a quasi-image format and leveraging pre-trained model weights based on images effectively enhance the model's classification ability. Additionally, the intermediate local selection component helped the model allocate more attention to the middle 10 seconds of 50-second-long samples, consistent with the behavior of human experts during annotation. The removal of this component resulted in performance loss, with non-significant difference (0.226 vs. 0.216, p = 0.413). Statistical analysis was performed using the Wilcoxon rank-sum test. The statistical sample comprised-patient level KLD results within the development cohort, involving a total of 1,063 patients. On the online testing dataset, VIPEEGNet exhibited similar performance. Specifically, the KLD values for VIPEEGNet, VIPEEGNet without central selection, VIPEEGNet without pre-training, and VIPEEGNet without EEG-to-image were 0.255, 0.271, 0.301, and 0.326, respectively. These results underscore the importance of each component in the overall performance of VIPEEGNet.

\subsection{T-SNE visualization}
T-SNE visualization was employed to reduce the dimensionality of the model's output and visualize the model's classification of different EEG sample categories\cite{van2008visualizing}. The t-SNE results resembled a starfish shape, as shown in Figure \ref{fig:fig4}b. Using the line connecting the seizure area to the “other” area as a dividing line, GPD and GRDA were located above the line, while LPD and LRDA were below. This indicates that the model has a high ability to distinguish between lateralized and generalized events. Moreover, GPD and LPD were closer to the seizure side, while GRDA and LRDA were closer to the “other” side, suggesting that VIPEEGNet tends to associate GPD and LPD more with seizures. This visualization provides an intuitive understanding of how the model differentiates between various EEG patterns and supports the findings from the classification performance metrics.

\section{Discussion}
This study aims to address the automatic classification of harmful brain activities, with the key challenge lying in utilizing the advances of pre-trained weights from image-trained algorithms to enhance model classification performance. An automated classifier, i.e. VIPEEGNET, which enables the direct adaptation of ImageNet pre-trained vision models for one-dimensional EEG signal analysis was developed. The research validated that employing the EEG-to-image conversion component and pre-trained models can improve performance by 21.5\% and 16.3\%, respectively. In practical applications, the parameter usage of VIPEEGNet is significantly less than that of existing SOTA algorithms, demonstrating its feasibility for real-world deployment.

Previous researchers have found that different experts exhibit variability when annotating different EEG rhythms and periodic patterns, with reliability only at a moderate level (overall pairwise agreement percentage of 52\%)\cite{jing2023interrater}. The variability in expert annotations can be attributed to several factors. First, EEG signals are complex and may be affected by various factors, such as individual differences in brain anatomy and physiology, changes in recording conditions, and artifacts generated by eye movements, muscle activity, etc., which increase the difficulty of accurate annotation\cite{dan2024szcore}. Second, the annotation of EEG rhythms and periodic patterns requires extensive expertise and experience. Different experts may have different understandings and judgments of the same EEG signal, leading to inconsistent annotations. Therefore, it is crucial to examine the generalization performance of different architectures. Ideally, the performance of deep learning algorithms should be consistent across datasets\cite{handa2024software}.

T-SNE analysis indicates that the model's extracted features for LPD and GPD samples are closer to seizure, while LRDA and GRDA are closer to the “other” side. This suggests that VIPEEGNet tends to associate GPD and LPD more with epileptic seizures, consistent with previous research findings\cite{zhou2025investigation}. Furthermore, a study based on 4,772 participants revealed that LPDs, regardless of frequency, are most strongly associated with epileptic seizures\cite{ruiz2017association}. The association between GPDs and LRDA and epileptic seizures is frequency-dependent. For LPDs and GPDs, the higher the pattern frequency, the greater the risk of epileptic seizures. Typically, diffuse periodic EEG patterns often indicate acute encephalopathy, while non-reactive bilateral rhythmic patterns may suggest epilepticus\cite{gelisse2025rhythmic}.

Previous studies have shown that pre-trained neural networks can effectively extract deep EEG features. Researchers have demonstrated the effectiveness and feasibility of pre-trained models in EEG classification tasks\cite{liu2023eeg, hammour2024scalp, klein2025flexible}. By varying the scale of pre-training data, model performance generally improves with increasing data volume. However, compared to natural language and image data, acquiring EEG data presents significant challenges\cite{jiang2024large}. Additionally, EEG data annotation typically requires substantial effort from domain experts, leading to a lack of sufficient EEG datasets for training baseline models for subsequent fine-tuning on downstream tasks.

When tested on the online testing cohort, numerous top-performing algorithms utilized a combination of raw EEG, spectrogram, and waveform plots to train multi-view models. While this approach can enhance generalization to some extent, it also increases model complexity and opacity, posing deployment challenges\cite{dong2025multi}. Moreover, according to the Data Processing Inequality, practical data processing steps such as time-frequency transformation and waveform plotting inevitably result in information loss. Researchers still choose to convert raw EEG into spectrogram and waveform plots to match the input requirements of pre-trained models\cite{nogay2021detection}. This is because transfer learning models must adhere to the dimensions of pre-trained models and cannot easily modify the neural architecture to address classification problems in other datasets\cite{yuen20243d}. To address the challenges of applying image-trained models to EEG classification tasks, VIPEEGNet employs a data-driven approach to automatically learn image representations of raw EEG signals.

\subsection{Strengths and Limitations}
VIPEEGNet combines pre-trained models in the field of computer vision, such as EfficientNetV2-B3, with multi-channel one-dimensional EEG signal analysis. By designing a universal EEG-to-image conversion module, 1D EEG signals are converted into image representations. The feature extraction ability of the ImageNet pre-trained model is directly utilized to improve the robustness of the model. In external validation, compared with the champion algorithm, the model parameter quantity was reduced by more than 96\%, meeting the real-time monitoring needs of clinical practice. The EEG-to-image component can be adapted to other pre-trained visual models, providing a foundation for future extensions. Visually displaying the distribution of model features conforms to medical cognition and enhances clinical credibility. At the same time, algorithms have certain limitations. In reality, patients' EEG may be more complex, indicating the need for more data and the accumulation of more EEG patterns\cite{ruijter2022treating}. The six identified EEG patterns are highly relevant to epileptic and seizure disorders, encephalitis, meningitis, and impaired consciousness. However, it remains to be explored whether VIPEEGNet can serve as a universal algorithmic basis for EEG pattern classification in other diseases, such as sleep disorders, metabolic and neurodegenerative diseases.

\section{Conclusion}
In summary, this study offers new insights into auto-detection and classification of seizures and other harmful brain activities. The VIPEEGNet, trained on EEG signals from intensive care patients, demonstrates that transfer learning from vision models enables efficient, clinically applicable EEG classification. It offers a deployable solution for real-time brain monitoring, which potentially may expand access to expert-level EEG interpretation in resource-limited settings and facilitate precise treatment for epilepsy. 

\section*{Acknowledgments}
This study was funded by Scientific Research Innovation Capability Support Project for Young Faculty (ZYGXQNJSKYCXNLZCXM-H15), National Science Fund for Excellent Overseas Scholars (0401260011), National Natural Science Foundation of China (82472098, 32300704), Tianjin Natural Science Foundation-Outstanding Youth Project (24JCJQJC00250), Major Science and Technology Special Projects and Engineering-Major Project of National Key Laboratories (24ZXZSSS00510), National Key Technologies Research and Development Program (2021YFF1200602), Non-profit Central Research Institute Fund of Chinese Academy of Medical Sciences (2024-JKCS-16)

\bibliographystyle{unsrt}  
\bibliography{references}  

\end{document}